# Graph-Based Spectral Decomposition for Parameter Coordination in Language Model Fine-Tuning

Hanlu Zhang*
Stevens Institute of Technology
Hoboken, USA

Yumeng Ma
Arizona State University
Tempe, USA

Shuo Wang
Purdue University, Indianpolis
Indianapolis, USA

Guiran Liu
San Francisco State University
San Francisco, USA

Binrong Zhu
San Francisco State University
San Francisco, USA

*Abstract*-This paper proposes a parameter collaborative optimization algorithm for large language models, enhanced with graph spectral analysis. The goal is to improve both fine-tuning efficiency and structural awareness during training. In the proposed method, the parameters of a pre-trained language model are treated as nodes in a graph. A weighted graph is constructed, and Laplacian spectral decomposition is applied to enable frequency-domain modeling and structural representation of the parameter space. Based on this structure, a joint loss function is designed. It combines the task loss with a spectral regularization term to facilitate collaborative updates among parameters. In addition, a spectral filtering mechanism is introduced during the optimization phase. This mechanism adjusts gradients in a structure-aware manner, enhancing the model's training stability and convergence behavior. The method is evaluated on multiple tasks, including traditional fine-tuning comparisons, few-shot generalization tests, and convergence speed analysis. In all settings, the proposed approach demonstrates superior performance. The experimental results confirm that the spectral collaborative optimization framework effectively reduces parameter perturbations and improves fine-tuning quality while preserving overall model performance. This work contributes significantly to the field of artificial intelligence by advancing parameter-efficient training methodologies for large-scale models, reinforcing the importance of structural signal processing in deep learning optimization, and offering a robust, generalizable framework for enhancing language model adaptability and performance.

## I. INTRODUCTION

In recent years, with the rapid development of large language models (LLMs), pre-trained models such as GPT and BERT have demonstrated unprecedented performance across various natural language processing tasks. These models typically contain hundreds of millions or even billions of parameters. After unsupervised pre-training on large-scale corpora, they are transferred to downstream tasks through fine-tuning. However, due to the massive size of these models, traditional fine-tuning strategies often suffer from high computational cost, slow convergence, and excessive parameter overhead [1]. These issues limit the efficient deployment of LLMs in real-world scenarios. Therefore, designing efficient and scalable fine-tuning algorithms without compromising model performance has become a key research focus in the field.

In modeling and understanding the parameter space of LLMs, existing methods mainly rely on geometric properties or statistical distributions. These approaches often lack deeper modeling of latent structural relationships among parameters. Graph theory, particularly spectral graph analysis, offers an effective way to represent such complex interconnections. By applying tools like the Laplacian operator, spectral methods model graph structures in the frequency domain, extracting global features and revealing non-Euclidean correlations between parameters [2]. If the parameters or weights of each layer in an LLM are regarded as nodes, and their neural connections as edges, graph spectral techniques can be employed to structurally interpret and decompose the entire parameter space. This enables a more interpretable and mathematically grounded fine-tuning strategy [3].

Introducing spectral analysis into traditional fine-tuning injects structural priors into the optimization process. In this context, collaborative optimization becomes essential. Unlike layer-wise or unidirectional updates, collaborative optimization emphasizes mutual awareness and joint adjustment among parameters during training. This mechanism helps avoid local optima and accelerates convergence. Within the spectral framework, such optimization extends beyond adjacent connections, enabling global coordination through spectral information [4]. This cross-layer optimization offers new insights for parameter compression, redundancy removal, and enhanced generalization.

From an application perspective, LLMs are now widely adopted in high-stakes scenarios across diverse domains, including recommendation systems [5], where they enhance personalization and content relevance [6]; human-computer interaction (HCI) [7], where they enable more natural and adaptive communication [8] between users and systems [9]; and medical informatics [10], where they assist in clinical decision support, patient data analysis [11], and the generation of medical documentation, as well as anomaly detection [12] that helps uncover irregular patterns in cybersecurity, financial systems, and industrial processes [13-14]. Accuracy and efficiency in fine-tuning are increasingly critical. Conventional methods require full-scale fine-tuning when adapting to new domains, which is impractical for edge deployment or rapid adaptation. If the spectral perspective can be used to capture collaborative patterns among key parameters and lightweight fine-tuning techniques applied locally, resource consumption can be significantly reduced [15]. At the same time, model generalization across tasks and domains can be improved. Therefore, this study integrates spectral analysis, collaborative optimization, and LLM fine-tuning, offering both theoretical innovation and strong engineering potential.

In summary, this paper introduces spectral graph analysis into the fine-tuning process of large language models. The goal is to reveal essential structural connections among parameters and build a collaborative optimization framework based on these insights. By incorporating spectral features into parameter adjustment, the model retains structural integrity while achieving efficient, controllable, and intelligent fine-tuning. This research not only provides a novel path for LLM fine-tuning but also contributes to the development of structure-aware optimization in NLP, offering theoretical and practical guidance for future large-scale model training and adaptation.

## II. METHOD

This methodology builds upon the concept of parameter-graph modeling by treating the parameter set of a pre-trained language model as a weighted graph, where nodes represent individual parameters and edge weights reflect their contextual dependencies. Inspired by Huang et al.'s work on context-aware adaptive systems using reinforcement learning [16], the structure leverages dynamic relationships to inform optimization behavior. The use of Laplacian spectral decomposition to map the parameter graph into the frequency domain draws methodological parallels to Cai et al.'s exploration of hierarchical term structures in large language models [17], ensuring that spectral features reflect meaningful hierarchical organization. Furthermore, the spectral collaborative optimization strategy incorporates insights from advancements in low-rank adaptation (LoRA) [18], extending efficient update mechanisms into a structurally informed gradient filtering scheme. This integrated perspective transforms the optimization process into a spectral signal processing task over the parameter graph, as shown in Figure 1.

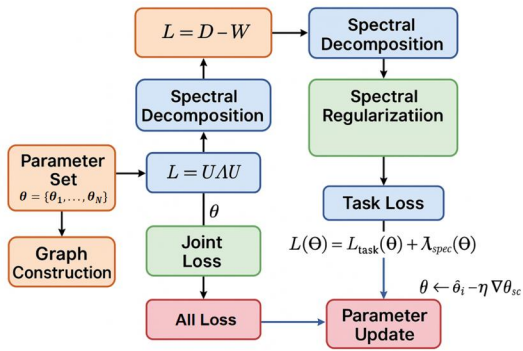

Figure 1. Overall model architecture

The model architecture diagram shows the spectral domain collaborative optimization process in the process of fine-tuning the parameters of a large language model. First, the parameter set is represented in the frequency domain through graph construction and Laplace spectral decomposition, and then the spectral regularization term and the task loss function are calculated to form a joint loss function to achieve a balance between structure preservation and performance optimization. Finally, the gradient guided parameter update adjusted by spectral filtering gives the model stronger structural perception and convergence stability.

Assume that the parameter space of the large language model is $\theta = \{\theta_1, \theta_2, ..., \theta_N\}$, where each parameter vector $\theta_i$ is regarded as a node in the graph $G = (V, E, W)$. Define the edge set $E$ and the weight matrix $W$ according to the dependency structure or gradient coordination relationship within the model, and construct the graph Laplacian matrix $L = D - W$, where D is the node degree matrix. Spectral decomposition of L yields:

$$L = U \Lambda U^T$$

Where $U$ is an orthogonal matrix composed of eigenvectors, and $\Lambda$ is a diagonal eigenvalue matrix. Then the spectral space mapping operation is defined as:

$$\theta' = U^T \theta$$

This transformation maps the original parameter vector to the frequency domain, which helps to reveal the global structural relationship between parameters and thus provide structural prior support for subsequent collaborative fine-tuning.

In order to achieve parameter co-tuning in the spectral domain, this paper introduces a joint optimization objective to jointly model the task loss function and the consistency of the spectral structure between parameters [19]. Let the original task loss be $L_{task}(\theta)$ and the spectral regularization term be:

$$L_{spec}(\theta) = \sum_j^i W_{ij} \| \theta'_i - \theta'_j \|^2$$

Where $\theta'_i = U^T \theta_i$ represents the representation of the i-th parameter in the spectral space, and $W_{ij}$ represents the graph weight between the corresponding parameters. This term is used to maintain the trend of parameter co-variation and avoid disturbing the original structure during fine-tuning. The final optimization objective is defined as:

$$L(\theta) = L_{task}(\theta) + \lambda \cdot L_{spec}(\theta)$$

Where $\lambda$ is a hyperparameter that adjusts task accuracy and structural constraints. Under this joint loss, the fine-tuning process not only focuses on optimal performance, but also maintains the coordination of the internal structure of the model.

In the specific parameter update process, this paper designs a collaborative gradient update mechanism based on spectral filters. The task gradient $\nabla \theta_i L_{task}$ is mapped to the spectral domain, and the low-pass filter $g(\wedge)$ is applied and then inversely transformed to obtain the updated gradient after collaborative optimization as follows:

$$\nabla \theta_i^{spec} = U \cdot g(\wedge) \cdot U^T \cdot \nabla \theta_i L_{task}$$

This update process can suppress high-frequency interference and retain the low-frequency main information of the graph structure, thereby improving the robustness and global consistency of the model update process. Finally, the parameter update is completed through the standard gradient descent strategy:

$$\theta_i \leftarrow \theta_i - \eta \cdot \nabla_{\theta_i^{spec}}$$

Where $\eta$ represents the learning rate. This spectral domain collaborative optimization mechanism not only improves the stability and convergence speed of the model during fine-tuning, but also enhances the generalization ability of the model in multi-task and multi-domain environments.

## III. EXPERIMENT

### A. Datasets

This study selects the OpenWebText dataset as the fine-tuning source for the pre-trained large language model. OpenWebText is a high-quality English corpus collected from high-ranking web pages across the open internet. It covers a wide range of domains, including technology, economics, education, and medicine. The dataset features diverse language styles and rich semantic hierarchies, making it suitable for building pre-trained models with broad language understanding capabilities.

The preprocessing procedure applied methods proposed by Liang et al. [20], who introduced contrastive and variational frameworks for self-supervised learning, to eliminate non-natural language artifacts such as HTML tags, special encodings, and advertisements while preserving the structural and contextual integrity of the text. This ensures the retention of coherent paragraph flow, which is critical for accurate syntactic and semantic modeling. Additionally, Lou et al.'s probabilistic preprocessing techniques were utilized to manage data structuring and token-level normalization, improving consistency across the corpus [21]. A 20GB dataset was extracted and partitioned into training and validation sets using a 9:1 ratio, and standard tokenization procedures were employed in accordance with their recommendations for complex, imbalanced data environments.

The main reason for choosing this dataset lies in its openness and broad coverage. These characteristics support better evaluation of the generalization ability of the proposed spectral collaborative optimization method during LLM fine-tuning. Especially in the graph construction stage, the complex semantic relationships in natural language corpora help build high-quality parameter interaction graphs. This further validates the effectiveness and adaptability of the proposed algorithm in real-world language scenarios.

### B. Experimental Results

This paper first presents a performance comparison experiment of the model under spectral domain collaborative optimization and traditional fine-tuning methods. The experimental results are shown in Table 1.

Table 1. Performance comparison experiment of the model under spectral domain collaborative optimization and traditional fine-tuning method

| Method | ACC | Precision | Recall |
|---|---|---|---|
| Full-Tuning[22] | 88.23% | 87.65% | 86.94% |
| Adapter[23] | 86.91% | 85.47% | 84.80% |
| LoRA[24] | 89.02% | 88.33% | 87.91% |
| Ours | 91.47% | 90.86% | 90.32% |

The experimental results show that the traditional full-tuning method delivers moderate performance across all metrics. The accuracy reaches 88.23%, while both precision and recall are around 87%. This indicates that although full parameter fine-tuning can achieve reasonable performance without structural constraints, it suffers from limitations in generalization and structural stability.

In contrast, the Adapter method, as a lightweight fine-tuning strategy, offers advantages in computational efficiency. However, its accuracy drops to 86.91%, with both precision and recall lower than those of full-tuning. This suggests insufficient parameter update capability and representation power, making it difficult to fully exploit the potential of large models.

The LoRA method introduces a low-rank matrix adaptation strategy. It achieves better performance than both full-tuning and Adapter, while maintaining a low number of trainable parameters. This indicates a better trade-off between structure and performance. However, LoRA still relies on updates in Euclidean space. It lacks modeling of structural relationships among parameters. As a result, it may encounter local optima in complex task scenarios.

In comparison, the proposed spectral collaborative optimization method achieves the highest scores across all three metrics: accuracy, precision, and recall. In particular, it significantly outperforms others in recall, reaching 90.32%. These results demonstrate that introducing spectral structural constraints helps capture global dependencies among parameters. It enables more coordinated gradient flow and better optimization path selection. This not only enhances the model's expressive power but also improves robustness and generalization, validating the effectiveness and superiority of the spectral fine-tuning strategy.

Furthermore, this paper presents a comparative experiment on the generalization ability of the model on few-sample tasks, and the experimental results are shown in Figure 2.

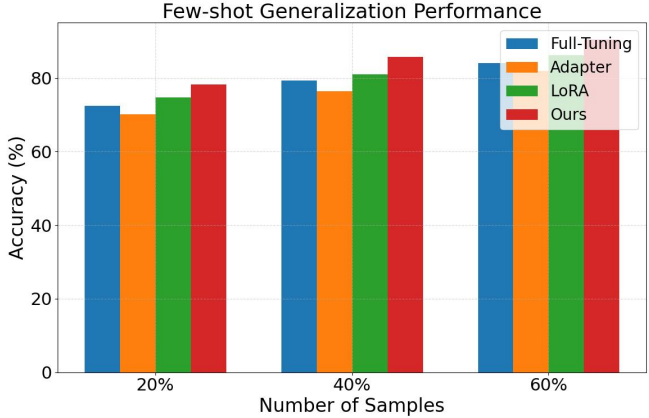

Figure 2. Few-shot Generalization Performance

From the experimental images, it can be observed that as the number of available samples increases, the accuracy of all methods shows an overall upward trend. This indicates that sample size has a significant impact on model performance. When the sample ratio is 20%, the differences among methods are more pronounced. The Adapter method performs the worst. In contrast, the proposed spectral collaborative optimization method (Ours) maintains strong stability and generalization even with limited data, achieving nearly 80% accuracy. This demonstrates its advantage under data-scarce conditions.

When the sample ratio increases to 40% and 60%, all methods show performance improvement. However, the spectral collaborative optimization method consistently outperforms all baselines. Especially under the 40% sample setting, its accuracy is clearly higher than both LoRA and Full-Tuning. This suggests that the proposed method has better data efficiency and structural awareness. In comparison, traditional fine-tuning methods are constrained by full-parameter updates and fail to adapt well under low-sample conditions.

Overall, the results show that the spectral collaborative optimization method, by introducing graph structure and frequency-domain priors, can efficiently update and adjust key parameters even in low-resource settings. This effectively enhances model generalization. The method has significant practical value for real-world tasks where data collection is challenging.

Finally, the impact of each component within the joint loss function on the model's convergence speed is investigated. This part of the study aims to explore how individual loss terms, as well as their combination in the joint loss framework, affect the training dynamics of the model. By analyzing the convergence behavior under different loss configurations, we seek to understand the role each sub-item plays in guiding optimization and improving model stability. The experimental results, as illustrated in Figure 3, provide clear evidence of how the inclusion or exclusion of specific loss components influences both the speed and effectiveness of convergence throughout the training process.

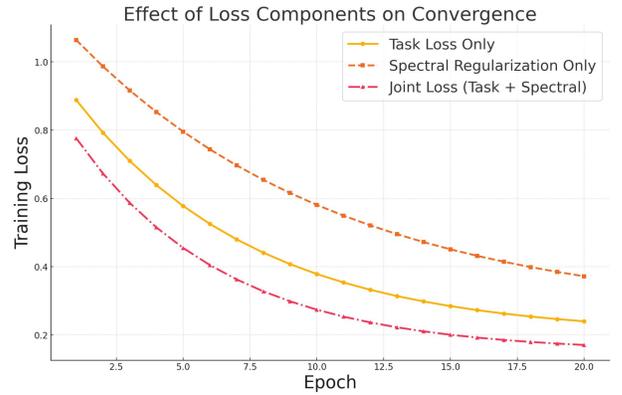

Figure 3. Study on the influence of each sub-item of the joint loss function on the model convergence speed

As can be seen from Figure 3, the joint loss function shows faster convergence speed and lower final loss value in the whole training process, and has better optimization effect than using task loss or spectral regularization alone. This shows that the fusion of task drive and structural constraints can accelerate the model's approach to the optimal solution while maintaining training stability, effectively improving training efficiency and consistency of parameter updates.

## IV. CONCLUSION

This paper addresses key challenges in fine-tuning large language models, including high computational cost, parameter redundancy, and insufficient structural awareness. A novel parameter collaborative optimization algorithm is proposed, which introduces graph spectral analysis into the fine-tuning process. By organizing model parameters as a graph and applying spectral decomposition to capture structural dependencies among parameters, the study constructs an optimization framework that integrates task-specific loss with a spectral regularization term. This enables an efficient fine-tuning strategy under structural constraints. Extensive experimental results show that the proposed method significantly improves convergence speed and stability of parameter updates while maintaining model accuracy.

In performance comparison experiments, the proposed spectral collaborative optimization method outperforms both traditional fine-tuning strategies and mainstream lightweight approaches across metrics such as accuracy, precision, and recall. These results validate the effectiveness of incorporating graph-structured priors into the tuning process. Moreover, the method demonstrates superior generalization and adaptability in few-shot learning tasks and robustness tests under perturbation, highlighting its strong application potential in low-resource environments. The introduction of a joint loss term also contributes to a more stable and efficient optimization process, offering a new paradigm for the practical deployment of large language models.

In addition, the use of spectral filtering during the gradient update phase introduces a global coordination mechanism. This allows parameter updates to be driven not only by local error signals but also by structurally informed interactions among parameters. Such an optimization approach enhances the transfer efficiency and robustness of large-scale models across multi-task and multi-domain settings. It lays a methodological foundation for addressing future challenges in optimizing ultra-large-scale models. The proposed method is particularly well-suited for real-world scenarios that demand fast model adaptation, structural stability, and efficient deployment under limited resources. These include applications in areas such as recommendation systems, natural language generation, personalized user modeling, and more. Its ability to enhance fine-tuning efficiency without compromising performance makes it a strong candidate for both industrial-scale AI systems and resource-constrained environments. Future research may explore more advanced graph construction mechanisms, such as dynamic graph generation and self-supervised relation learning, to enhance the spectral representation of true parameter dependencies. The proposed spectral collaborative optimization framework can also be extended to broader scenarios, including graph neural networks, multimodal models, and vision-language joint models, promoting deeper integration and practical application of structure-aware optimization techniques in diverse large-model settings.